\pdfoutput=1
\documentclass[conference]{IEEEtran}
\IEEEoverridecommandlockouts

\usepackage[utf8]{inputenc}
\usepackage{times}

\usepackage{amsmath}
\usepackage{amssymb}
\usepackage{amsthm}

\usepackage{graphicx}
\usepackage{booktabs}
\usepackage{multirow}

\usepackage{algorithm}
\usepackage{algorithmic}

\usepackage{url}
\usepackage[hidelinks]{hyperref}
\usepackage[small]{caption}
\usepackage{soul}
\usepackage{tcolorbox}
\usepackage{enumitem}

\newtheorem{example}{Example}

\newtheorem{definition}{Definition}

\title{\textsf{GAPL}: Grounded Action-effect Policy Learning\\for LLM-Based Trajectory Planning}

\author{
  \IEEEauthorblockN{
    Zhihong Cui\textsuperscript{1,\dag,*},
    Hengyu Liu\textsuperscript{2,\dag},
    Zhangkai Wu\textsuperscript{3},
    Yushuai Li\textsuperscript{2},\\
    Tianyi Li\textsuperscript{2},
    Peiyuan Guan\textsuperscript{1,4},
    Amir Taherkordi\textsuperscript{1},
    Tor Skeie\textsuperscript{1,5}
  }
  \IEEEauthorblockA{
    \textsuperscript{1}Department of Informatics, University of Oslo, Oslo, Norway\\
    \textsuperscript{2}Department of Computer Science, Aalborg University, Aalborg, Denmark\\
    \textsuperscript{3}University of Sydney, Sydney, Australia\\
    \textsuperscript{4}School of Software, Nanjing University of Information Science and Technology, Nanjing, China\\
    \textsuperscript{5}Simula Research Laboratory, Oslo, Norway\\
    \{zhihongc, peiyuang, amirhost, tskeie\}@ifi.uio.no,\quad
    \{heli, yusli, tianyi\}@cs.aau.dk,\quad
    amasawawoo@gmail.com
  }
  \thanks{\textsuperscript{\dag}These authors contributed equally.}
  \thanks{\textsuperscript{*}Corresponding author: Zhihong Cui (zhihongc@ifi.uio.no).}
}

\begin{document}

\maketitle

\begin{abstract}
Trajectory planning for autonomous driving requires both high-level reasoning and precise low-level control. Large Language Models (LLMs) offer semantic-rich planning capabilities, however, their application is limited by hallucinated reasoning, poor grounding in environment dynamics, and limited numerical precision in control. We propose \textsf{GAPL} (\textbf{G}rounded \textbf{A}ction-effect \textbf{P}olicy \textbf{L}earning), a unified framework that integrates LLM-based effect estimation, simulation-based effect grounding, and policy optimization into a closed-loop system. \textsf{GAPL} consists of three modules: (1) an LLM-based Effect Evaluator for structured multi-dimensional action-effect estimation; (2) a Simulation-based Effect Grounder that predicts dynamics-consistent effects from simulator rollouts; and (3) an Effect-Aware Decision Maker that grounds LLM effect estimates against simulation via a distiller to guide Proximal Policy Optimization (PPO)-based policy learning. Experiments on four Highway-env scenarios demonstrate that \textsf{GAPL} consistently outperforms baselines, achieving average reductions of \{0.76, 0.86, 2.00\} in collision rate, average displacement error (ADE), and final displacement error (FDE), and an average reward gain of 1.44.
\end{abstract}


\section{Introduction}
Trajectory planning is a core challenge in autonomous driving, requiring safe and efficient path generation in complex and dynamic environments~\cite{dong2025enhancing}. Traditional methods \cite{pourkeshavarz2024cadet,lin2024safety} are often scenario-specific, limiting adaptability to diverse traffic conditions. Recent advances in LLMs, with strong reasoning and contextual understanding~\cite{lan2024traj}, enable semantic-rich planning and more adaptive driving behaviors~\cite{fu2024drive}. However, LLM-based planning remains challenging due to hallucinated reasoning~\cite{jin2024position,sun2025understanding}, poor grounding in environment dynamics~\cite{yang2025comprehensive,yin2025grounding}, and limited numerical precision for control~\cite{feng2025numerical,wu2025scot}, hindering reliable execution.

\begin{figure}
  \centering
  \includegraphics[width=0.95\linewidth]{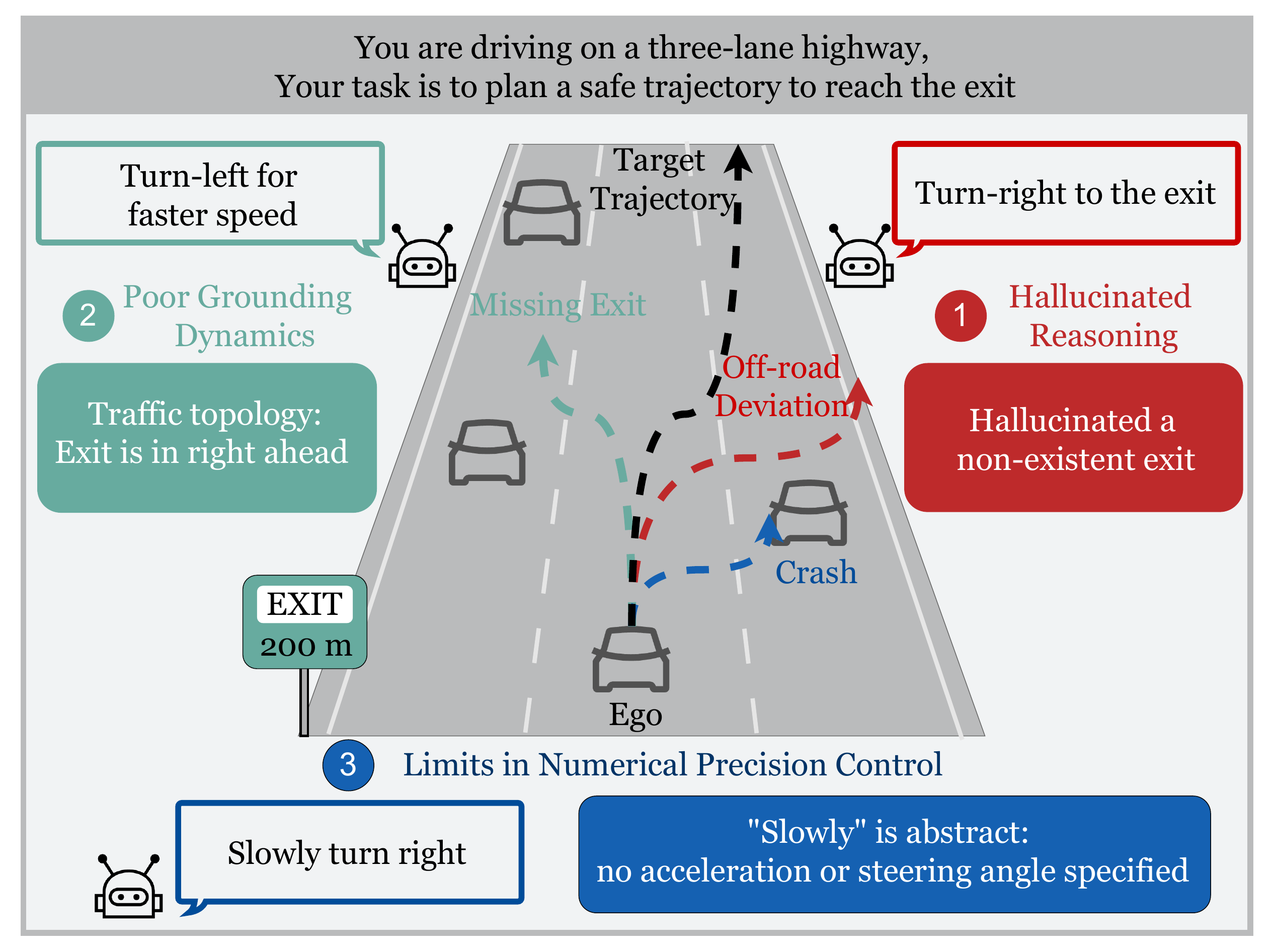}
  \caption{Limitations of LLM-based Trajectory Planning}
  \vspace{-5mm}
  \label{fig:exampleForLLMLimitation}
\end{figure}
\begin{example}
Consider an ego vehicle on a three-lane highway, as shown in Fig.~\ref{fig:exampleForLLMLimitation}, the target trajectory (black) should guide the vehicle to smoothly merge right and safely reach the designated highway exit. However, the LLM generates three flawed candidate trajectories (red, green, and blue): 1) Due to hallucination, the vehicle hallucinates a non-existent exit, causing it to drift out of the lane. 2) The green trajectory shows an illegal left lane change despite exit is located on the right. This reflects a lack of grounding to environment dynamics, such as road topology. 3) Lacking precise control signals, the blue trajectory follows a vague instruction (“Slowly turn right”), which may cause the vehicle to stall or behave abnormally, increasing the risk of a side collision.
\end{example}

These observations suggest that reliable LLM-based trajectory planning requires \textbf{grounding semantic reasoning in environment dynamics}. To this end, we aim to develop a unified framework that \textbf{leverages simulation feedback to constrain LLM predictions and translates the aligned decisions into executable control policies}.
However, existing LLM-based trajectory planning methods have not achieved such grounding. Broadly, they fall into two categories. (1) \textit{LLM-driven planners}~\cite{ashwani2024cause} directly generate actions from language descriptions, relying on detailed prompts to encode domain knowledge and reduce hallucination~\cite{jin2024position}. However, language alone cannot capture real-time environment dynamics~\cite{carta2023grounding}, such as vehicle interactions, road geometry constraints, and physical feasibility remain opaque to the LLM~\cite{chen2025next}, leading to plans that are semantically plausible but physically inconsistent~\cite{Dung2025Navigating}. (2) \textit{LLM-assisted planners}~\cite{gkountouras2024language} mitigate this by delegating trajectory execution to external planners while restricting the LLM to high-level decisions. Although this improves numerical precision~\cite{feng2025numerical}, the LLM itself remains ungrounded: it operates in an open-loop manner without feedback from environment interaction~\cite{hao2025rl}. As a result, errors from hallucinated reasoning~\cite{sun2025understanding} or domain knowledge gaps propagate through the planning pipeline, giving rise to three challenges:

\noindent \textit{\textbf{C1: How to constrain hallucinated planning outputs of LLMs?}}
A major cause of hallucinations in LLM-based planning is the lack of explicit structural action evaluation. Most existing methods~\cite{tang2025plan,gendron2025causal} rely on free-form language descriptions or implicit rewards, making it difficult for LLMs to explicitly assess the action effects of candidate actions. In autonomous driving, each action simultaneously affects multiple effect dimensions~\cite{dal2025joint}, such as safety risk, energy consumption, and driving comfort. However, these dimensions are rarely modeled in a structured manner during LLM-based planning, leading to unsafe or suboptimal actions. The challenge lies in how to introduce structured, multi-dimensional effect representations into LLM-based planning.

\noindent \textit{\textbf{C2: How to align language-based effect estimation with environment dynamics?}} Even with structured effect representations, LLM-derived estimates remain inconsistent with real environment dynamics~\cite{carta2023grounding}.
Domain-specific factors such as vehicle interactions, road geometry, and physical execution constraints are difficult to fully encode through language prompts alone~\cite{chen2025next}. Although fine-tuning LLMs on driving data~\cite{ma2024coevolving,zhai2024fine} could improve grounding, it is computationally expensive. Therefore, a key challenge is how to align language-based effect assessments with environment dynamics in a data-efficient manner.

\noindent \textbf{\textit{C3: How to ensure numerically precise and executable control decisions?}} While LLMs are well-suited for high-level reasoning, they are not designed to generate precise numerical control~\cite{requeima2024llm}, which is essential for autonomous driving. Even when high-level decisions are correct, imprecise control parameters such as position, velocity, and acceleration can lead to unsafe trajectories in real-world driving. This raises a critical challenge: how to translate calibrated action effects into executable control policies that satisfy the precision requirements of driving.

To address the above challenges, we propose \textsf{GAPL}, an LLM-based trajectory planning framework that integrates
structured effect estimation, simulation-grounded effect references, and precision policy learning within a unified
decision-making loop. The framework consists of three modules: an LLM-based Effect Evaluator, a Simulation-based Effect Grounder, and an Effect-Aware Decision Maker. \textbf{To address C1}, the LLM-based Effect Evaluator prompts the LLM to perform structured multi-dimensional effect estimation over actions and select candidate actions based on their predicted action effects (e.g., reward, crash risk, energy, and comfort). This structured effect evaluation constrains hallucinated planning outputs of LLMs in complex driving scenarios. \textbf{To address C2}, the Simulation-based Effect Grounder learns to predict dynamics-consistent trajectory-level action effects from simulator rollouts for candidate actions. By computing the same effect metrics as the LLM evaluator, it provides references that serve as alignment signals for subsequent calibration. \textbf{To address C3}, the Effect-Aware Decision Maker employs a grounding distiller that calibrates LLM-predicted effects toward the simulation-grounded estimates into a unified training signal, which is combined with environment rewards to guide PPO-based policy learning through environment interaction, enabling the agent to learn numerically precise and executable control strategies for trajectory planning.

Our main contributions are summarized as follows:
\begin{itemize} [itemsep=2pt, leftmargin=12pt]
    \item We identify three tightly coupled challenges in LLM-based trajectory planning: unreliable effect estimation under hallucination, misalignment with environment dynamics, and insufficient numerical precision for control.

    \item We propose \textsf{GAPL}, a unified framework that integrates structured effect estimation, simulation-based effect grounding, and policy optimization into a closed-loop decision-making pipeline for trajectory planning.

\item Experiments on four Highway-env scenarios demonstrate that \textsf{GAPL} consistently outperforms baselines, achieving average reductions of \{0.76, 0.86, 2.00\} in collision rate, ADE, and FDE, and an average reward gain of 1.44.
\end{itemize}

\vspace{+2mm}
\section{Related Work}
\vspace{+1mm}
\subsection{LLM-Driven Planner}
Recent studies~\cite{lin2025improving,song2023llm} utilize LLMs directly for trajectory planning. SayCan~\cite{Ahn2022can} and Reflexion~\cite{shinn2023reflexion} prompt LLMs with natural language state descriptions to generate high-level actions or policies. While these methods demonstrate impressive reasoning capabilities, they often suffer from hallucinations and lack grounding in physical dynamics. Code-as-Policies~\cite{liang2023code} and SayCanPay~\cite{hazra2024saycanpay} introduce structured prompts and tool-augmented reasoning to enhance execution reliability.
However, LLMs are trained on text corpora and lack grounding in real-world dynamics, leading to unreliable planning.

\subsection{LLM-Assisted Planner}
Recent methods~\cite{gkountouras2024language,kambhampati2024position} confine LLMs to high-level decision-making, while offloading effect estimation to separate transition models. For example, Plan-R1~\cite{tang2025plan} evaluates candidate actions through rollout-based simulation for action selection, and Cartographer~\cite{gendron2025causal} further refines simulated effects using predefined graph structures. While these approaches improve execution feasibility, they remain largely decoupled from policy optimization and provide indirect supervision for learning effective control strategies.  Other methods~\cite{dong2025enhancing,chen2025next} introduce natural language feedback to refine LLM decisions. However, such feedback is typically sparse and weakly structured, which limits its effectiveness for stable policy learning. In contrast, our method explicitly formulates trajectory planning as a system-level learning problem, requiring coordinated treatment of effect reasoning, environment-grounded alignment, and control policy optimization within a closed-loop planning framework.

\section{Preliminary}
\label{sec:preliminaries}

\begin{definition}[\textbf{Action-Effect Environment Model}]
\label{def:env_model}
We consider an action-effect environment model for trajectory planning defined as a tuple
\(
\mathcal{M} = (\mathcal{S}, \mathcal{A}, \mathcal{T}, \mathcal{G}, \mathcal{O}),
\)
where $\mathcal{S}$ is the state space, $\mathcal{A}$ the action space, $\mathcal{T}$ the environment transition function, $\mathcal{G}$ a set of instantaneous effect functions, and $\mathcal{O}$ represents the multi-dimensional action-effect space capturing the downstream consequences of executing actions. An \textbf{action effect} quantifies the anticipated impact of an action across multiple evaluation dimensions (e.g., task progress, safety, efficiency, comfort)~\cite{liu2025diffugc,zhou2021multiobjrl}, providing richer supervisory signals than scalar rewards alone.
\end{definition}

In this setting, $\mathcal{S}$ represents the state space of the ego and surrounding vehicles, including positions, velocities, and orientations.
$\mathcal{A}$ consists of discrete high-level actions such as \textit{turn-left}, \textit{turn-right}, \textit{accelerate}, \textit{IDLE}, and \textit{decelerate}.
The transition function $\mathcal{T}$ models environment dynamics and is realized by a simulator that evolves the state as: $ s_{t+1} = \mathcal{T}(s_t, a_t)$. $\mathcal{G} = \{g_r, g_p, g_e, g_c\}$ are instantaneous effect functions that evaluate step-level reward, crash risk, energy consumption, and comfort.

\begin{definition}[\textbf{Trajectory-level Action Effect}]
\label{def:traj_effect}
Given a state $s_t$ and an action $a \in \mathcal{A}$, the trajectory-level action effect of executing $a$ at $s_t$ is defined as the aggregated effects over a finite horizon $L$:
\begin{equation}
\begin{aligned}
 & \boldsymbol{o}_t(s_t, a) =(o_{R,t},o_{P,t},o_{E,t},o_{C,t}) = \\
& (
 \sum_{\ell=0}^{L-1} g_r(s_{t+\ell}, a_{t+\ell}),
\sum_{\ell=0}^{L-1} g_p(s_{t+\ell}, a_{t+\ell}), \\
& \sum_{\ell=0}^{L-1} g_e(s_{t+\ell}, a_{t+\ell}),
\sum_{\ell=0}^{L-1} g_c(s_{t+\ell}, a_{t+\ell})),
\end{aligned}
\end{equation}
where $o_{R,t}$, $o_{P,t}$, $o_{E,t}$, and $o_{C,t}$ represent cumulative action effects on reward, crash risk, energy consumption, and comfort. We distinguish two sources of action effects: (i) \textbf{semantic effects} $\boldsymbol{o}^{\text{\tiny LLM}}_t$ estimated by an LLM via linguistic reasoning, and (ii) \textbf{grounded effects} $\boldsymbol{o}^{\text{\tiny GRO}}_t$ obtained from simulator rollouts reflecting dynamics-consistent action effects.
\end{definition}

\noindent
\textbf{Effect Grounding.} We define \textit{grounding} as the process of aligning semantic effect estimates with dynamics-consistent simulator observations. This alignment reduces the discrepancy between LLM-predicted effects and environment dynamics, enabling more reliable effect-aware policy learning.

\subsection{Problem Statement}

We formulate trajectory planning as an \emph{effect-aware policy learning} problem. At each time step $t$, the agent observes the current driving state $s_t \in \mathcal{S}$. An LLM proposes a candidate action set $\mathcal{A}_t \subseteq \mathcal{A}$, and each candidate action $a \in \mathcal{A}_t$ is associated with an estimated trajectory-level action-effect vector $\boldsymbol{o}_t(s_t,a) \in \mathbb{R}^4$, capturing cumulative reward, crash risk, energy consumption, and driving comfort over a finite horizon. The objective is to learn a policy $\pi_\eta(a_t \mid s_t, \mathcal{A}_t)$ that selects an executable action $a_t \in \mathcal{A}_t$ at each step, such that the resulting state-action sequence forms safe, efficient, and comfortable trajectories in dynamic driving environments.

\section{GAPL}
\label{sec:method}

\subsection{Framework}
\begin{figure*}[htp]
  \centering
  \includegraphics[width=\linewidth]{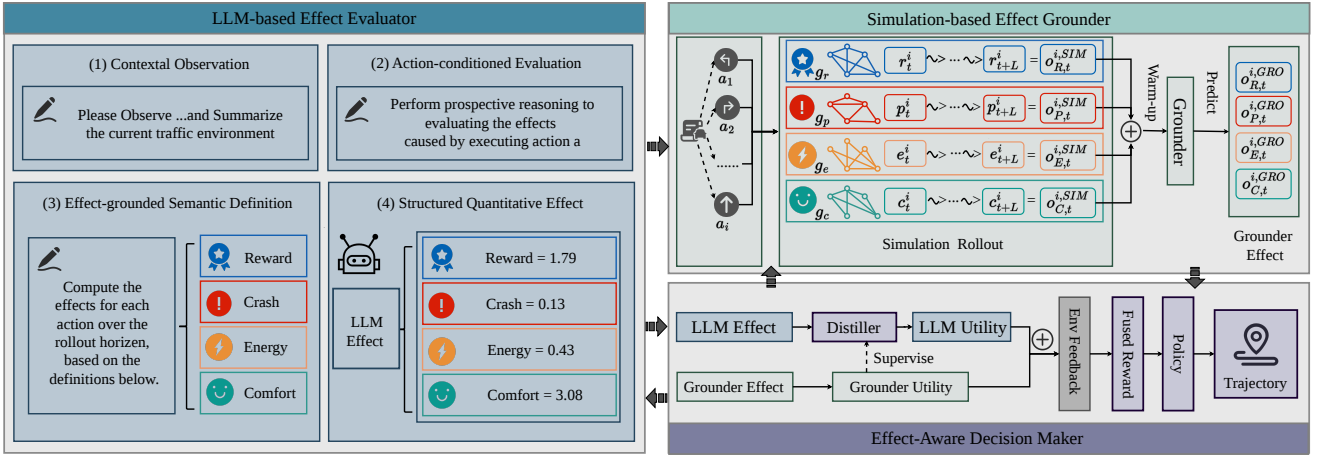}
  \vspace{-4mm}
  \caption{The overview of GAPL framework}
  \label{fig:framework}
  \vspace{-4mm}
\end{figure*}

\textsf{GAPL}, as illustrated in Fig.~\ref{fig:framework}, consists of three modules:
an LLM-based Effect Evaluator, a Simulation-based Effect Grounder, and an Effect-Aware Decision Maker.
Given the current state $s_t$, the Effect Evaluator prompts the LLM with structured action-conditioned
prompts to perform effect estimation over possible actions and to select $n$ candidate actions
$\{a^i\}_{i=1}^n$, each accompanied by multi-dimensional numerical action-effect estimates, including expected
reward, crash risk, energy consumption, and comfort, denoted as
$\boldsymbol{o}^{\text{\tiny LLM}}_t(s_t, a^i)$. Next, the selected candidate actions are evaluated by the Simulation-based Effect Grounder, which is trained to approximate dynamics-consistent action effects obtained from simulator rollouts. By computing the same effect metrics $\boldsymbol{o}^{\text{\tiny GRO}}_t(s_t, a^i)$ as
the LLM evaluator, it provides dynamics-consistent references that supervise the subsequent
calibration of the LLM effects. Finally, the Effect-Aware Decision Maker employs a grounding distiller that calibrates LLM-predicted
effects toward the simulation-grounded estimates, producing a grounded utility $\tilde{o}^{i}_{t}$ that is combined with environment
rewards to guide PPO-based policy learning through environment interaction. Sequential execution of policy-selected actions under environment dynamics induces trajectories, realizing closed-loop trajectory planning.

\subsection{LLM-based Effect Evaluator}
Most existing LLM-based planners rely primarily on reward-centric or heuristic signals
\cite{Ahn2022can,ashwani2024cause}, which is insufficient for real-world driving, where each action simultaneously affects safety, efficiency, energy consumption, and comfort. To address this issue, we introduce an LLM-based Effect Evaluator that prompts the LLM to estimate a structured action-effect vector for each action. Specifically, it outputs quantitative action-effect estimation over four dimensions (e.g., reward, crash risk, energy, and comfort).

As illustrated in the left part of Fig.~\ref{fig:framework}, the structured multi-dimensional effect prompt guides LLM to perform action-effect estimation through the following mechanisms:
\setlength{\parsep}{0pt}
\begin{itemize}[itemsep=2pt, leftmargin=10pt]
 \item \textbf{Contextual observation.} The prompt instructs the model to summarize the current traffic and ego history trajectories (e.g., \textit{``Please Observe ....and Summarize ...''}), providing context for effect assessment.
    \item \textbf{Action-conditioned evaluation.} For each candidate action, the prompt asks the model to assess \textit{what would likely happen if this action were executed}, encouraging prospective effect estimation over future trajectories.
    \item \textbf{Effect-grounded semantic definitions.} Each effect dimension~\cite{qian2024adaptraj} is described using grounded language criteria:reward reflects task progress; crash risk corresponds to the probability of collision; energy consumption is associated with the magnitude of control effort (e.g., acceleration); and comfort is inversely related to abrupt changes in acceleration. These definitions constrain the LLM to produce numerically interpretable and dynamics-consistent estimates.
    \item \textbf{Structured quantitative outputs.} The LLM is required to return numerical estimates in a structured format:
    \begin{equation}
    \begin{aligned}
     \boldsymbol{o}^{i,\text{\tiny LLM}}_{t}(s_t, a^i) & =  \text{LLM}(\text{\textit{prompt}}(s_t, a^i)) \\
     & = 
    \left(o^{i,\text{\tiny LLM}}_{R,t},\; o^{i,\text{\tiny LLM}}_{P,t},\; o^{i,\text{\tiny LLM}}_{E,t},\; o^{i,\text{\tiny LLM}}_{C,t}\right),
    \end{aligned}
    \end{equation}
where the LLM is accessed via an external API and remains fixed throughout training~\cite{hu2025efficient}, and
$i \in \{1,\dots,n\}$ indexes the candidate actions.
$o^{i,\text{\tiny LLM}}_{R,t}$, $o^{i,\text{\tiny LLM}}_{P,t}$, $o^{i,\text{\tiny LLM}}_{E,t}$, and
$o^{i,\text{\tiny LLM}}_{C,t}$ denote the predicted $L$-step cumulative reward, crash risk, energy, and comfort, respectively. Here, the LLM itself serves as effect functions $\mathcal{G}=\{g_r,g_p,g_e,g_c\}$ that directly maps $(s_t, a^i)$ to trajectory-level action-effect estimates.
\end{itemize}

By enforcing explicit multi-dimensional action-effect estimation, the LLM is encouraged to explicitly assess the long-term consequences of candidate actions, thereby reducing hallucinated decisions. We use GPT-4.1-nano as default LLM backbone. The detailed prompt template is provided in Appendix~\ref{sec:Appendix_B}, and a computational overhead analysis is provided in Appendix~\ref{sec:Appendix_E}. Additional results with different LLM backbones are provided in Appendix~\ref{sec:Appendix_LLMbb}.

\subsection{Simulation-based Effect Grounder}
Although the LLM can provide action-effect estimates from linguistic priors, these predictions may deviate from true environment dynamics~\cite{Dung2025Navigating}. To address this, we introduce a Simulation-based Effect Grounder, which learns to predict trajectory-level action effects.

\noindent \textbf{Effect Grounder Model.} As shown in the upper-right part of Fig.~\ref{fig:framework}, the grounder takes the current state $s_t$ and $n$ candidate actions $\{a^i\}_{i=1}^n$ as input, and predicts trajectory-level action effects for each action:
\begin{equation}
\begin{aligned}
\boldsymbol{o}^{i,\text{\tiny GRO}}_{t}(s_t, a^i) &= \text{GRO}_\phi(s_t, a^i)\\
& = \left(o^{i,\text{\tiny GRO}}_{R,t},\; o^{i,\text{\tiny GRO}}_{P,t},\; o^{i,\text{\tiny GRO}}_{E,t},\; o^{i,\text{\tiny GRO}}_{C,t}\right),  
\end{aligned}
\end{equation}
where $\text{GRO}_\phi$ is a multi-headed Transformer~\cite{zhu2023towards} that jointly predicts four effect metrics over $L$-step horizon, including expected reward $o^{i,\text{\tiny GRO}}_{R,t}$, crash risk $o^{i,\text{\tiny GRO}}_{P,t}$, energy $o^{i,\text{\tiny GRO}}_{E,t}$, and comfort $o^{i,\text{\tiny GRO}}_{C,t}$.

\noindent \textbf{Effect Target Generation via Simulator Rollouts.}
To train the effect grounder, we generate supervision targets based on the environment transition function $\mathcal{T}$~\cite{liao2024cdstraj} with stochastic perturbations:
\begin{equation}
\begin{aligned}
\boldsymbol{o}^{i,\text{\tiny SIM}}_{t}(s_t, a^i) & = \left(o^{i,\text{\tiny SIM}}_{R,t},\; o^{i,\text{\tiny SIM}}_{P,t},\; o^{i,\text{\tiny SIM}}_{E,t},\; o^{i,\text{\tiny SIM}}_{C,t}\right),  \\
& \sim \mathcal{T}(\cdot \mid s^{i}_{t+\ell}, a^{i}; \xi_{t+\ell}), 
\end{aligned}
\end{equation}
where each candidate action $a^i$ is repeatedly executed to produce an action-conditioned rollout trajectory, yielding a state sequence $\{s^{i}_{t+\ell}\}_{\ell=0}^L$. Stochastic disturbances $\xi_{t+\ell}$ affect the rollout only through state transitions and are implicitly reflected in the sampled trajectories.

Along each rollout trajectory, the simulator computes the trajectory-level action effect by aggregating the instantaneous effect functions in different forms:
\begin{equation}
\label{equGROfour}
    \begin{aligned}
    o^{i,\text{\tiny SIM}}_{R,t} & =\sum_{\ell=0}^{L-1} \gamma^\ell \, g_r(s^{i}_{t+\ell}, a^i), \\
    o^{i,\text{\tiny SIM}}_{P,t} & = 1 - \prod_{\ell=0}^{L-1} \Big(1 - g_p(s^{i}_{t+\ell}, a^i)\Big),\\
 o^{i,\text{\tiny SIM}}_{E,t} &= \sum_{\ell=0}^{L-1} g_e(s^{i}_{t+\ell}, a^i),\\
 o^{i,\text{\tiny SIM}}_{C,t} & = \sum_{\ell=0}^{L-1} g_c(s^{i}_{t+\ell}, a^i),
    \end{aligned}
\end{equation}
where $\gamma \in (0,1]$ is discount factor. Different aggregation forms reflect the semantics of each effect dimension. Reward is discounted to emphasize near-term progress, crash risk models the probability of at least one collision during the rollout, and energy and comfort are accumulated to measure total control effort and motion smoothness. We instantiate the instantaneous effect functions $\mathcal{G}=\{g_r,g_p,g_e,g_c\}$ as:
\begin{equation}
\label{eq:g_defs}
\begin{aligned}
g_r(s^{i}_{t+\ell}, a^i) &= r^{i}_{t+\ell},\\
g_p(s^{i}_{t+\ell}, a^i) &= p^{i}_{t+\ell},\\
g_e(s_{t+\ell}^i, a^i) & = \|u_{t+\ell}^i\|^2,\\
g_c(s_{t+\ell}^i, a^i) & = \|(\dot v_{t+\ell}^i - \dot v_{t+\ell-1}^i)\|^2,
\end{aligned}
\end{equation}
where $r^{i}_{t+\ell}$ and $p^{i}_{t+\ell}$ are reward and crash-risk returned by the simulator. $u^{i}_{t+\ell}$ denotes the low-level continuous control input (e.g., acceleration and steering) induced by executing the high-level action $a^i$, and $v_t$ and $\dot v_t$ denote vehicle velocity and corresponding acceleration.

\noindent \noindent \textbf{Warm-up Pre-training.}
We pre-train the effect grounder using data collected via a uniform random policy, ensuring diverse state-action coverage before policy optimization.
Formally, the grounder $\text{GRO}_\phi(s_t, a^i)$ is trained to regress to simulator-derived trajectory-level action-effect targets $\boldsymbol{o}^{i,\text{\tiny SIM}}_{t}(s_t, a^i)$ by minimizing a multi-task regression loss:
\begin{equation}
\mathcal{L}_{\text{\tiny GRO}}(\phi)
=\mathbb{E}\left[
\big\|
\text{GRO}_\phi(s_t,a^i)-\boldsymbol{o}^{i,\text{\tiny SIM}}_t
\big\|_2^2
\right].
\end{equation}
After warm-up, the grounder is fixed and used to provide fast and stable action-effect estimates for subsequent training stages. Training curves of the four effect dimensions are reported in our anonymized code repository.

\subsection{Effect-Aware Decision Maker}
Given LLM-predicted effects and grounder-predicted effects for candidate actions, the LLM estimates are semantically rich but may be miscalibrated to environment dynamics. We therefore introduce an Effect-Aware Decision Maker that uses the dynamics-consistent grounder to calibrate the LLM effects into a grounded utility, which is combined with environment rewards as shaped signals for PPO-based policy learning.

\noindent \textbf{Grounding Distiller Module.}
The grounder, trained on simulator rollouts, yields dynamics-consistent effect estimates that we treat as a reference. We use a distiller $\mathcal{D}_\theta$~\cite{ha2023scaling} to calibrate the LLM's semantic effects toward this reference, while the already-grounded grounder effects are scored directly by a utility function $\psi(\cdot)$:
\begin{equation}
\label{equ:sharedDistiller}
    \begin{aligned}
    \tilde{o}^{i,\text{\tiny LLM}}_{t} & = \alpha \cdot \mathcal{D}_\theta\!\left(\boldsymbol{o}^{i,\text{\tiny LLM}}_t\right) + (1-\alpha)\cdot \text{Nor}\!\left(r_t^{\text{env}}\right), \\
    \tilde{o}^{i,\text{\tiny GRO}}_{t} & = \alpha \cdot \psi\!\left(\boldsymbol{o}^{i,\text{\tiny GRO}}_t\right) + (1-\alpha)\cdot \text{Nor}\!\left(r_t^{\text{env}}\right),
    \end{aligned}
\end{equation}
where $\psi(\boldsymbol{o}) = \text{Nor}(o_{R}) - \omega_P\, o_{P} - \omega_E\,\text{Nor}(o_{E}) + \omega_C\,\text{Nor}(o_{C})$ aggregates the four effect dimensions into a scalar utility, $\text{Nor}(\cdot)$ is normalization, the residual $(1-\alpha)\text{Nor}(r_t^{\text{env}})$ provides environment supervision during early training, and $\alpha \in [0,1]$ balances the two terms. The grounder effects, being dynamics-consistent, are scored directly by $\psi$ to give the \emph{grounder utility} $\tilde{o}^{i,\text{\tiny GRO}}_t$, whereas the LLM effects pass through $\mathcal{D}_\theta$ to give the \emph{LLM utility} $\tilde{o}^{i,\text{\tiny LLM}}_t$; $\mathcal{D}_\theta$ is trained (below) to reproduce the grounder-grade utility from the LLM's semantic estimates.

\noindent
\textbf{Reward Shaping for Policy Optimization.}
To translate the LLM and grounder utilities into actionable learning signals for policy optimization, we propose a shaped reward by combining both calibrated utilities with the environment reward.
\begin{equation}
\label{equ:shapedreward}
\hat{r}_t = 
\omega_{\text{\tiny env}} \cdot r_t^{\text{env}} +
\omega_{\text{\tiny LLM}} \cdot \tilde{o}^{\text{\tiny LLM}}_{t} +
\omega_{\text{\tiny GRO}} \cdot \tilde{o}^{\text{\tiny GRO}}_{t},
\end{equation}
where $r_t^{\text{env}}$ is the reward returned by the simulator after executing action $a_t$, and $\omega_{\text{\tiny env}}, \omega_{\text{\tiny LLM}}, \omega_{\text{\tiny GRO}} \in [0,1]$ control the contribution of each term. A sensitivity study on different weight configurations is provided in Appendix~\ref{sec:Appendix_F}. By retaining the environment reward, policy optimization remains aligned with the true task objective, while the distilled utility terms provide dense auxiliary guidance for decision making.

\noindent
\textbf{Policy Execution and Trajectory Generation.}
We learn a stochastic policy $\pi_\eta(a_t \mid s_t)$ over the environment’s native action space. At each time step, an action is sampled from the policy and executed in the environment:
\begin{equation}
a_t \sim \pi_\eta(\cdot \mid s_t), \quad s_{t+1} \sim \mathcal{T}(s_t, a_t).
\end{equation}
The dynamic interactions induce a trajectory
$\tau = \{(s_t, a_t, \hat r_t, s_{t+1})\}_{t=0}^{T}$.
Policy parameters $\eta$ are optimized using trajectories collected on-policy under the shaped reward.

\subsection{Joint Training}
The joint training objective combines policy optimization and distiller, formulated as:
\begin{equation}
\mathcal{L}_{\text{total}} = \lambda_{\text{\tiny PPO}}\mathcal{L}_{\text{PPO}}(\eta)
+ \lambda_{\text{\tiny Dis}} \cdot \mathcal{L}_{\text{Dis}}(\theta),
\end{equation}
where $\lambda_{\text{\tiny PPO}}$ and $\lambda_{\text{\tiny Dis}}$ balance policy learning and distiller.

\noindent
\textbf{Policy Training.}
We adopt the clipped PPO algorithm for policy learning.
The loss function is defined as:
\begin{equation}
\begin{aligned}
& \mathcal{L}_{\text{PPO}}(\eta)
= \mathbb{E}_t\!\left[
\min\!\left(
\rho_t A_t,\;
\text{clip}(\rho_t, 1 \pm \epsilon) A_t
\right)
\right], \\
& A_t = \sum_{\ell=0}^{L-1} (\gamma \lambda)^\ell \delta_{t+\ell}, \quad
\delta_t = \hat{r}_t + \gamma V_\eta(s_{t+1}) - V_\eta(s_t),
\end{aligned}
\end{equation}
where $A_t$ is the generalized advantage estimate computed from the shaped reward $\hat{r}_t$, $\delta_t$ is the temporal-difference residual, $\rho_t$ the policy ratio, $\epsilon$ the clipping threshold, $\gamma$ the discount factor, $\lambda$ the GAE parameter, and $V_\eta(s)$ the value function.

\noindent
\textbf{Distiller Training.}
The distiller $\mathcal{D}_\theta$ is trained to \emph{ground} the LLM's semantic effects by regressing them to the grounder-derived utility:
\begin{equation}
\label{equ:distillerloss}
\mathcal{L}_{\text{Dis}}(\theta) = \mathbb{E}_t \Big[
\big( \mathcal{D}_\theta(\boldsymbol{o}^{i,\text{\tiny LLM}}_t) - \text{sg}(\psi(\boldsymbol{o}^{i,\text{\tiny GRO}}_t)) \big)^2
\Big],
\end{equation}
where $\text{sg}(\cdot)$ is the stop-gradient operator and the grounder is frozen after warm-up, providing a stable dynamics-grounded target $\psi(\boldsymbol{o}^{i,\text{\tiny GRO}}_t)$. This trains $\mathcal{D}_\theta$ to correct hallucinated or miscalibrated LLM estimates toward dynamics-consistent values, thereby realizing grounding. Because the target derives from the simulation-grounded grounder rather than from $\mathcal{D}_\theta(\boldsymbol{o}^{i,\text{\tiny GRO}}_t)$ itself, the objective cannot be minimized by a trivial constant.

\begin{table*}[!htbp]
\centering
\caption{Performance Comparison on Four Driving Scenarios Compared With Baselines (mean$\pm$std).}
\label{tab:overallresults}
\small
\setlength{\tabcolsep}{4pt}
\renewcommand{\arraystretch}{1.1}
\resizebox{\linewidth}{!}{%
\begin{tabular}{c|c|cc|cc|ccc|c|c}
\toprule
\textbf{Scenario} & \textbf{Metrics}
    & \textbf{SayCan}
    & \textbf{Reflexion}
    & \textbf{CaDet}
    & \textbf{FUSION}
    & \textbf{Plan-R1}
    & \textbf{Cartographer}
    & \textbf{LLMCWM}
    & \textbf{\textsf{GAPL}}
    & \textbf{p-value} \\
\midrule
\multirow{4}{*}{Hig.}
& CR ($\downarrow$,\%) & 9.36 ± 1.24 & 8.89 ± 1.11 & 6.72 ± 1.05 & 4.18 ± 0.21 & 3.94 ± 0.29 & 3.62 ± 0.26  & 3.11 ± 0.32 & \textbf{2.18 ± 0.11} & $\textbf{1.13}\mathrm{\textbf{e}}{\textbf{-3}}$\\
& ADE ($\downarrow$) & 10.52 ± 2.31 & 9.63 ± 2.07 & 10.26 ± 2.42 & 8.83 ± 2.51 & 8.57 ± 1.93 & 7.23 ± 1.64 & 7.04 ± 1.51 & \textbf{6.04 ± 0.89} & $\textbf{1.47}\mathrm{\textbf{e}}{\textbf{-3}}$\\
& FDE ($\downarrow$) & 15.23 ± 3.02 & 13.45 ± 2.73 & 14.15 ± 3.12 & 12.08 ± 3.23 & 11.52 ± 2.85 & 10.86 ± 2.6  & 9.61 ± 2.33 & \textbf{6.45 ± 1.04} &$\textbf{1.22}\mathrm{\textbf{e}}{\textbf{-4}}$ \\
& Reward ($\uparrow$) & 6.28 ± 1.81 & 6.93 ± 2.07 & 8.45 ± 2.45 & 10.83 ± 2.11 & 11.57 ± 2.09 & 12.59 ± 1.83  & 13.61 ± 2.32 & \textbf{16.78 ± 1.61} & $\textbf{1.15}\mathrm{\textbf{e}}{\textbf{-4}}$ \\
\midrule

\multirow{4}{*}{Mer.}
& CR ($\downarrow$,\%) & 9.52 ± 1.21 & 9.13 ± 1.17 & 7.45 ± 0.95 & 4.97 ± 0.67 & 4.57 ± 0.59 & 4.12 ± 0.53  & 3.58 ± 0.52 & \textbf{2.72 ± 0.38} & $\textbf{3.16}\mathrm{\textbf{e}}{\textbf{-3}}$ \\
& ADE ($\downarrow$) & 9.18 ± 1.82 & 8.79 ± 1.50 & 9.33 ± 2.07 & 7.91 ± 2.16 & 8.08 ± 1.45 & 7.46 ± 1.33  & 7.22 ± 1.13 & \textbf{6.65 ± 1.01} & $\textbf{5.53}\mathrm{\textbf{e}}{\textbf{-3}}$\\
& FDE ($\downarrow$) & 13.52 ± 2.66 & 12.97 ± 2.15 & 13.27 ± 2.43 & 11.88 ± 2.84 & 11.55 ± 2.28 & 10.42 ± 1.92  & 9.35 ± 1.82 & \textbf{7.86 ± 1.10} & $\textbf{4.29}\mathrm{\textbf{e}}{\textbf{-3}}$\\
& Reward ($\uparrow$) & 3.34 ± 0.96 & 3.79 ± 1.18 & 4.29 ± 0.83 & 5.01 ± 0.76 & 5.59 ± 0.81 & 6.12 ± 0.78 & 6.68 ± 0.72 & \textbf{7.55 ± 0.68} & $\textbf{3.13}\mathrm{\textbf{e}}{\textbf{-3}}$\\
\midrule

\multirow{4}{*}{Rou.}
& CR ($\downarrow$,\%) & 10.59 ± 1.18 & 10.27 ± 1.17 & 9.88 ± 1.16 & 6.78 ± 0.80 & 6.32 ± 0.62 & 5.89 ± 0.60  & 5.38 ± 0.52 & \textbf{4.74 ± 0.41} & $\textbf{1.06}\mathrm{\textbf{e}}{\textbf{-3}}$\\
& ADE ($\downarrow$) & 11.27 ± 2.61 & 10.64 ± 2.23 & 11.47 ± 2.53 & 9.78 ± 2.81 & 9.34 ± 2.17 & 8.27 ± 1.56 & 7.95 ± 1.88 & \textbf{7.22 ± 1.04} & $\textbf{6.29}\mathrm{\textbf{e}}{\textbf{-3}}$\\
& FDE ($\downarrow$) & 15.99 ± 3.42 & 14.71 ± 3.07 & 15.07 ± 3.25 & 13.09 ± 3.54 & 12.26 ± 2.92 & 11.44 ± 2.65 & 10.19 ± 2.34 & \textbf{9.13 ± 1.06} & $\textbf{7.72}\mathrm{\textbf{e}}{\textbf{-3}}$\\
& Reward ($\uparrow$) & 2.78 ± 1.28 & 3.14 ± 1.22 & 3.68 ± 0.66 & 4.43 ± 0.62 & 5.02 ± 0.64 & 5.69 ± 0.57 & 6.27 ± 0.45 & \textbf{6.84 ± 0.41} & $\textbf{7.38}\mathrm{\textbf{e}}{\textbf{-3}}$\\
\midrule

\multirow{4}{*}{Int.}
& CR ($\downarrow$,\%) & 11.76 ± 1.62 & 11.25 ± 1.57 & 10.15 ± 1.52 & 7.35 ± 1.10 & 7.03 ± 0.89 & 6.24 ± 0.83  & 5.79 ± 0.88 & \textbf{5.18 ± 0.72} & $\textbf{6.53}\mathrm{\textbf{e}}{\textbf{-3}}$\\
& ADE ($\downarrow$) & 16.83 ± 4.25 & 14.72 ± 3.65 & 15.65 ± 4.03 & 12.24 ± 3.06  & 13.01 ± 2.62 & 11.88 ± 2.53  & 10.65 ± 2.23 & \textbf{9.52 ± 0.98} & $\textbf{3.13}\mathrm{\textbf{e}}{\textbf{-3}}$\\
& FDE ($\downarrow$) & 20.13 ± 4.71 & 18.37 ± 3.85 & 19.44 ± 4.51 & 15.62 ± 4.33  & 16.09 ± 3.44 & 14.23 ± 3.37 & 13.53 ± 3.07 & \textbf{11.23 ± 1.01} & $\textbf{1.18}\mathrm{\textbf{e}}{\textbf{-3}}$\\
& Reward ($\uparrow$) & 2.03 ± 0.91 & 2.46 ± 0.76 & 2.78 ± 0.50 & 3.53 ± 0.55 & 4.03 ± 0.62 & 4.52 ± 0.61  & 5.17 ± 0.52 & \textbf{6.33 ± 0.46} & $\textbf{1.21}\mathrm{\textbf{e}}{\textbf{-3}}$ \\
\bottomrule
\end{tabular}%
}
\vspace{-5mm}
\end{table*}

\section{Experiments}
\subsection{Experimental Settings}
\noindent \textbf{Datasets.} Experiments are conducted on the Highway-env simulator~\cite{highway-env} across four scenarios: \emph{highway}, \emph{merge}, \emph{roundabout}, and \emph{intersection}. Each episode runs for up to 1000 steps or ends upon collision or task completion.

\noindent \textbf{Baselines.} \textsf{GAPL} is compared with seven baselines across three categories: (1) \textbf{Classical planners}: CaDet~\cite{pourkeshavarz2024cadet}, FUSION~\cite{lin2024safety}; (2) \textbf{LLM-driven planners}: SayCan~\cite{Ahn2022can}, Reflexion~\cite{shinn2023reflexion}; (3) \textbf{LLM-assisted planners}: Plan-R1~\cite{tang2025plan}, Cartographer~\cite{gendron2025causal}, LLMCWM~\cite{gkountouras2024language}.

\noindent \textbf{Metrics.} We report four metrics: \textbf{Collision Rate (CR)}, the fraction of episodes terminating in a collision; \textbf{Average Displacement Error (ADE)} and \textbf{Final Displacement Error (FDE)}, the mean and final Euclidean deviation of the executed trajectory from the target; and cumulative \textbf{Reward} (Eq.~(\ref{equ:shapedreward})), a fused signal over environment feedback and LLM/simulation effects. Results are averaged over five independent runs. Detailed implementation and hyperparameter settings are provided in Appendix~\ref{sec:Appendix_A}.

\subsection{Comparison Performance}
We evaluate the trajectory planning performance of \textsf{GAPL} against seven representative baselines
across four driving scenarios: \emph{highway} (Hig.), \emph{merge} (Mer.), \emph{roundabout} (Rou.), and
\emph{intersection} (Int.). Performance is measured using CR, ADE, FDE, and cumulative reward, where lower
CR/ADE/FDE and higher reward indicate better planning quality. A significance test was conducted compared to the best baseline (LLMCWM). The results are reported in Table~\ref{tab:overallresults}.

As shown in Table~\ref{tab:overallresults}, \textsf{GAPL} consistently achieves superior performance across all scenarios and metrics. Compared with the strongest baseline (LLMCWM), \textsf{GAPL} reduces the collision rate by \{0.93, 0.86, 0.64, 0.61\}, ADE by \{1.00, 0.57, 0.73, 1.13\}, and FDE by \{3.16, 1.49, 1.06, 2.30\} in \emph{highway}, \emph{merge}, \emph{roundabout}, and \emph{intersection} scenarios, respectively, while achieving reward improvements of \{3.17, 0.87, 0.57, 1.16\}, with an average gain of 1.44 across scenarios. Besides, the significance experiment showed that the p-values are all less than 0.05, demonstrating the effectiveness of \textsf{GAPL}. These performance gains can be attributed to three factors. First, \textsf{GAPL} constrains LLM planning through structured multi-dimensional action-effect estimation,
which effectively mitigates hallucinated or logically unsafe actions.
Second, simulation-based effect grounder aligns LLM estimation with environment-grounded dynamics, injecting physically consistent feedback into the decision process. Finally, the grounding distiller calibrates the LLM's semantic predictions toward the grounder effects, providing stable and informative supervision for policy optimization.

Moreover, the comparison across different planner categories further supports this conclusion. Purely LLM-driven planners such as SayCan and Reflexion exhibit consistently poor performance across all scenarios, revealing the limitations of direct language-based action generation under complex traffic dynamics. In contrast, LLM-assisted planners that incorporate external simulators (e.g., LLMCWM, Plan-R1, and Cartographer) significantly outperform classical planners such as FUSION and CaDet, demonstrating the advantage of grounded dynamics in decision making. By jointly integrating structured semantic reasoning, simulation-based grounding, and closed-loop policy learning, \textsf{GAPL} further improves upon these baselines, achieving the best overall performance.

\subsection{Ablation Study}
To evaluate the contribution of each component in \textsf{GAPL}, we conduct an ablation study across four scenarios with three variants:
(1) \textbf{w/o LLM}, which uses only simulation-based effect estimation;
(2) \textbf{w/o SIM}, which relies solely on LLM-predicted effects; and
(3) \textbf{w/o Distiller}, which uses a standard PPO with raw environment rewards.

\begin{table}[!tbp]
\centering
\caption{Ablation Study Across Four Driving Scenarios.}
\label{tab:ablation}
\setlength{\tabcolsep}{1.2pt}
{
\fontsize{8.4pt}{10pt}\selectfont
\begin{tabular}{c|c|ccc|c}
\toprule
    & \textbf{Metric} & \textbf{w/o LLM} & \textbf{w/o SIM} & \textbf{w/o Distiller} & \textbf{All} \\
\midrule
\multirow{3}{*}{Hig.}
    & ADE ($\downarrow$) & 7.63 ± 2.11 & 9.10 ± 2.58 & 6.42 ± 1.93 & \textbf{6.04 ± 0.89} \\
    & FDE ($\downarrow$) & 8.14 ± 2.30 & 9.98 ± 2.79 & 6.81 ± 2.15 & \textbf{6.45 ± 1.04} \\
    & Reward ($\uparrow$) & 13.99 ± 1.73 & 10.84 ± 1.99 & 15.80 ± 1.69 & \textbf{16.78 ± 1.61} \\
\midrule
\multirow{3}{*}{Mer.}
    & ADE ($\downarrow$) & 8.24 ± 1.51 & 9.23 ± 1.70 & 7.22 ± 1.29 & \textbf{6.65 ± 1.01} \\
    & FDE ($\downarrow$) & 10.33 ± 2.37 & 12.08 ± 2.59 & 9.01 ± 2.21 & \textbf{7.86 ± 1.10} \\
    & Reward ($\uparrow$) & 5.61 ± 0.72 & 4.56 ± 0.78 & 6.22 ± 0.70 & \textbf{7.55 ± 0.68} \\
\midrule
\multirow{3}{*}{Rou.}
    & ADE ($\downarrow$) & 9.12 ± 0.24 & 10.04 ± 0.39 & 8.50 ± 0.12 & \textbf{7.22 ± 1.04} \\
    & FDE ($\downarrow$) & 11.50 ± 1.27 & 13.52 ± 1.31 & 10.15 ± 1.19 & \textbf{9.13 ± 1.06} \\
    & Reward ($\uparrow$) & 5.04 ± 0.18 & 4.29 ± 0.21 & 6.11 ± 0.17 & \textbf{6.84 ± 0.41} \\
\midrule
\multirow{3}{*}{Int.}
    & ADE ($\downarrow$) & 11.41 ± 1.36 & 13.26 ± 1.49 & 10.14 ± 1.30 & \textbf{9.52 ± 0.98} \\
    & FDE ($\downarrow$) & 13.56 ± 2.39 & 14.88 ± 2.50 & 12.82 ± 2.15 & \textbf{11.23 ± 1.01} \\
    & Reward ($\uparrow$) & 4.93 ± 0.31 & 3.14 ± 0.36 & 6.21 ± 0.29 & \textbf{6.33 ± 0.46} \\
\bottomrule
\end{tabular}
}
\vspace{-4mm}
\end{table}

As shown in Table~\ref{tab:ablation}, removing any component leads to consistent performance degradation, with simulation grounding being the most critical. Without simulation feedback, ADE and FDE increase by 3.05 and 3.95 on average, respectively, while reward decreases by 3.67, indicating that LLM-only estimation fails to capture dynamics. Removing the LLM also causes notable degradation, with average increases of 1.74 in ADE and 2.22 in FDE, and a reward drop of 1.98, suggesting reduced semantic generalization in complex scenes. The removal of the distiller causes the smallest overall drop, but still leads to noticeable drops in highly interactive scenarios such as \emph{merge} and \emph{intersection}.

Overall, the three components play complementary roles: the LLM provides structured semantic effect estimation, the simulator grounds predictions in physical dynamics, and the distiller integrates heterogeneous signals into stable policy learning. Their joint integration is essential for safe and generalizable trajectory planning.

\begin{figure}[t]
    \centering
    \includegraphics[width=\linewidth]{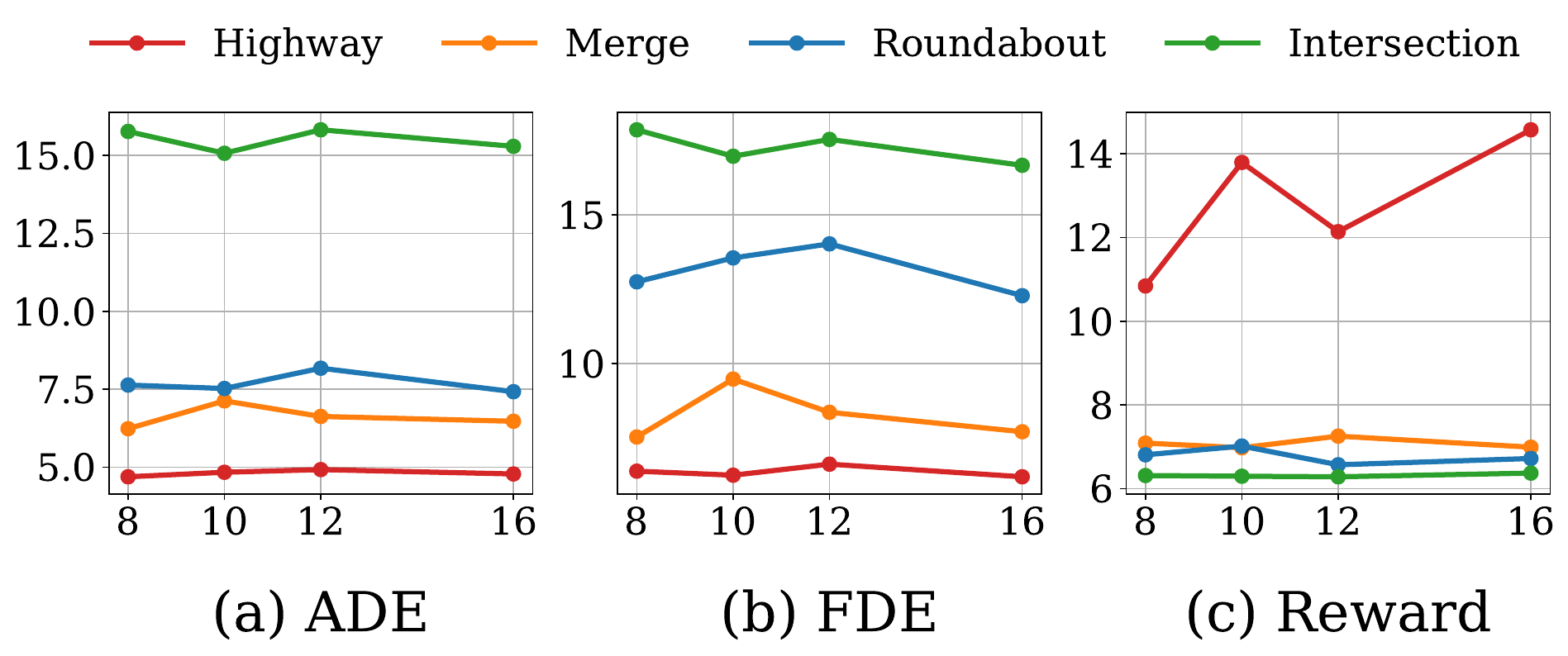}
    \vspace{-4mm}
    \caption{Planning performance under different rollout length}
    \label{fig:rollout}
    \vspace{-4mm}
\end{figure}

\begin{table}[htbp]
\centering
\caption{Performance Under Different Data Sizes for Grounder Warm-Up in the Intersection Scenario.}
\label{tab:warmup}
\setlength{\tabcolsep}{1.2pt}
{
\fontsize{8.4pt}{10pt}\selectfont
\begin{tabular}{c|ccc}
\toprule
\multicolumn{1}{c|}{Data size} & \textbf{ADE} ($\downarrow$)& \textbf{FDE} ($\downarrow$) & \textbf{Reward} ($\uparrow$) \\
\midrule
 0   & 13.88 ± 1.47 & 14.85 ± 2.31 & 3.11 ± 0.41 \\
 8k  & 12.41 ± 1.39 & 14.31 ± 2.22 & 4.03 ± 0.36 \\
 12k   & 11.01 ± 1.33 & 13.92 ± 2.14 & 4.54 ± 0.31 \\
 18k   & 10.85 ± 1.27 & 13.45 ± 2.10 & 5.67 ± 0.29 \\
 24k   & \textbf{9.52 ± 0.98} & \textbf{11.23 ± 1.01} & \textbf{6.33 ± 0.46} \\
\bottomrule
\end{tabular}
}
\vspace{-4mm}
\end{table}
\subsection{Sensitivity Study}
\noindent \textbf{Warm-up.}
As shown in Table~\ref{tab:warmup}, increasing warm-up data in the \emph{Intersection} scenario consistently improves performance. Without warm-up, unstable action–effect prediction causes large errors, while using 24k samples reduces ADE and FDE by 4.36 and 3.62, respectively, and increases cumulative reward by 3.22, indicating more reliable dynamics grounding before joint training.

\noindent \textbf{Rollout Length.}
Fig.~\ref{fig:rollout} evaluates the impact of rollout length $L$ on planning performance. \textsf{GAPL} achieves consistently low ADE/FDE and high reward for rollout lengths from 8 to 16 across all scenarios, with only mild variations observed. This robustness stems from aligning LLM semantic estimation with dynamics-consistent grounding, enabling horizon-invariant action evaluation. Additional sensitivity analysis on environment configurations (lane count and vehicle density) is reported in our anonymized code repository.

\begin{figure}[t]
    \centering
    \includegraphics[width=\linewidth]{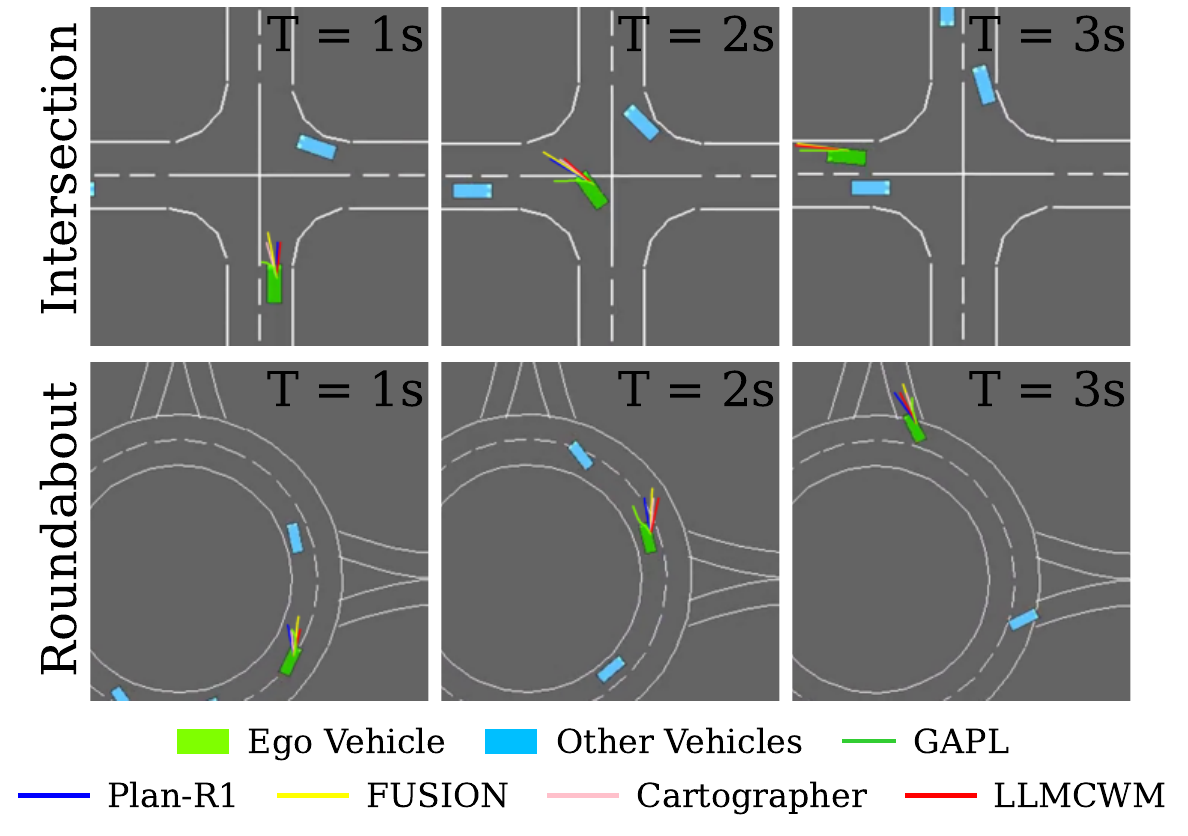}
    \vspace{-3mm}
    \caption{Case study of trajectory planning compared with baselines in intersection and roundabout scenarios}
    \label{fig:case}
    \vspace{-4mm}
\end{figure}

\subsection{Case Study}
To assess the trajectory planning capability of \textsf{GAPL}, we conduct a case study in \emph{intersection} and \emph{roundabout} scenarios. Trajectories are evaluated at discrete time steps ($T$ = 1s, 2s, and 3s). Four representative baselines are included for comparison, with all methods initialized from identical states.

In the \emph{intersection} scenario (Fig.~\ref{fig:case}, top), Plan-R1, Cartographer, and LLMCWM produce nearly straight trajectories at $T{=}1$s, indicating insufficient anticipation of the upcoming turn, which leads to abrupt turning at $T{=}2$s and persistent heading deviations at $T{=}3$s, increasing collision risk. In contrast, \textsf{GAPL} exhibits appropriate curvature already at $T{=}1$s, further stabilizes the trajectory at $T{=}2$s, and aligns well with the target lane by $T{=}3$s.
In the \emph{roundabout} scenario (Fig.~\ref{fig:case}, bottom), baselines show sharp trajectory corners across timestamps, reflecting unstable steering and degraded comfort, whereas \textsf{GAPL} maintains smooth curvature throughout, resulting in safer and more comfortable trajectories. Overall, these results demonstrate that \textsf{GAPL} consistently generates safer and smoother trajectories.
\section{Conclusion}
In this work, we identify three fundamental challenges that limit the effectiveness of LLM-based trajectory planning:
hallucinated reasoning, poor grounding in environmental dynamics, and limited numerical precision in control.
To address these limitations, we propose \textsf{GAPL}, a unified grounded action-effect policy learning framework that integrates LLM-based structured effect estimation, simulation-based effect grounding, and effect-aware policy optimization in a closed-loop manner. Extensive experiments across four challenging driving scenarios demonstrate that \textsf{GAPL} consistently outperforms state-of-the-art baselines, producing safer and smoother trajectories. These results suggest that tightly coupling semantic reasoning, dynamics-grounded simulation, and policy learning within a closed-loop framework is critical for reliable LLM-based trajectory planning. We also note limitations for future work: the grounder is frozen after warm-up and may degrade under policy-induced distribution shift (motivating online adaptation and grounder-guided LLM fine-tuning), is supervised by a simulator (leaving a sim-to-real gap), and models only four effect dimensions (extendable to, e.g., traffic-rule compliance).

\bibliographystyle{IEEEtran}
\bibliography{ijcai26}

\begin{thebibliography}{10}
\providecommand{\url}[1]{#1}
\csname url@samestyle\endcsname
\providecommand{\newblock}{\relax}
\providecommand{\bibinfo}[2]{#2}
\providecommand{\BIBentrySTDinterwordspacing}{\spaceskip=0pt\relax}
\providecommand{\BIBentryALTinterwordstretchfactor}{4}
\providecommand{\BIBentryALTinterwordspacing}{\spaceskip=\fontdimen2\font plus
\BIBentryALTinterwordstretchfactor\fontdimen3\font minus
  \fontdimen4\font\relax}
\providecommand{\BIBforeignlanguage}[2]{{%
\expandafter\ifx\csname l@#1\endcsname\relax
\typeout{** WARNING: IEEEtran.bst: No hyphenation pattern has been}%
\typeout{** loaded for the language `#1'. Using the pattern for}%
\typeout{** the default language instead.}%
\else
\language=\csname l@#1\endcsname
\fi
#2}}
\providecommand{\BIBdecl}{\relax}
\BIBdecl

\bibitem{dong2025enhancing}
Q.~Dong, T.~Wu, P.~Zeng, C.~zang, G.~Wan, and S.~Cui, ``Enhancing robot
  learning through cognitive reasoning trajectory optimization under unknown
  dynamics,'' \emph{IEEE Robotics and Automation Letters}, vol.~10, no.~6, pp.
  5401--5408, 2025.

\bibitem{pourkeshavarz2024cadet}
M.~Pourkeshavarz, J.~Zhang, and A.~Rasouli, ``Cadet: a causal disentanglement
  approach for robust trajectory prediction in autonomous driving,'' in
  \emph{Proceedings of the IEEE/CVF Conference on Computer Vision and Pattern
  Recognition}, 2024, pp. 14\,874--14\,884.

\bibitem{lin2024safety}
H.~Lin, W.~Ding, Z.~Liu, Y.~Niu, J.~Zhu, Y.~Niu, and D.~Zhao, ``Safety-aware
  causal representation for trustworthy offline reinforcement learning in
  autonomous driving,'' \emph{IEEE Robotics and Automation Letters}, vol.~9,
  no.~5, pp. 4639--4646, 2024.

\bibitem{lan2024traj}
Z.~Lan, H.~Li, L.~Liu, B.~Fan, Y.~Lv, Y.~Ren, and Z.~Cui, ``Traj-llm: A new
  exploration for empowering trajectory prediction with pre-trained large
  language models,'' \emph{arXiv preprint arXiv:2405.04909}, 2024.

\bibitem{fu2024drive}
D.~Fu, X.~Li, L.~Wen, M.~Dou, P.~Cai, B.~Shi, and Y.~Qiao, ``Drive like a
  human: Rethinking autonomous driving with large language models,'' in
  \emph{2024 IEEE/CVF Winter Conference on Applications of Computer Vision
  Workshops (WACVW)}.\hskip 1em plus 0.5em minus 0.4em\relax IEEE, 2024, pp.
  910--919.

\bibitem{jin2024position}
M.~Jin, Y.~Zhang, W.~Chen, K.~Zhang, Y.~Liang, B.~Yang, J.~Wang, S.~Pan, and
  Q.~Wen, ``Position: What can large language models tell us about time series
  analysis,'' in \emph{Proceedings of the International Conference on Machine
  Learning}, vol. 235, 2024, pp. 22\,260--22\,276.

\bibitem{sun2025understanding}
X.~Sun, J.~Liang, Y.~Wang, H.~Zhang, and D.~Zhao, ``Understanding visual detail
  hallucinations of large vision-language models,'' in \emph{Proceedings of the
  Thirty-Fourth International Joint Conference on Artificial Intelligence},
  2025, pp. 1900--1908.

\bibitem{yang2025comprehensive}
W.~Yang, L.~Some, M.~Bain, and B.~Kang, ``A comprehensive survey on integrating
  large language models with knowledge-based methods,'' \emph{Knowledge-Based
  Systems}, vol. 318, p. 113503, 2025.

\bibitem{yin2025grounding}
H.~Yin, H.~Qian, Y.~Shi, I.~Tsang, and Y.-S. Ong, ``Grounding open-domain
  knowledge from llms to real-world reinforcement learning tasks: A survey,''
  in \emph{Proceedings of the Thirty-Fourth International Joint Conference on
  Artificial Intelligence}, 2025, pp. 10\,797--10\,806.

\bibitem{feng2025numerical}
G.~Feng, K.~Yang, Y.~Gu, X.~Ai, S.~Luo, J.~Sun, D.~He, Z.~Li, and L.~Wang,
  ``How numerical precision affects arithmetical reasoning capabilities of
  llms,'' in \emph{Proceedings of Findings of the Association for Computational
  Linguistics}, 2025, pp. 46--85.

\bibitem{wu2025scot}
Z.~Wu, X.~Fan, H.~Wu, and L.~Cao, ``{SC}ot: {U}nifying {C}onsistency {M}odels
  and {R}ectified {F}lows via {S}traight-{C}onsistent {T}rajectories,'' in
  \emph{The Thirty-ninth Annual Conference on Neural Information Processing
  Systems}, 2025.

\bibitem{ashwani2024cause}
S.~Ashwani, K.~Hegde, N.~R. Mannuru, D.~S. Sengar, M.~Jindal, K.~C.~R. Kathala,
  D.~Banga, V.~Jain, and A.~Chadha, ``Cause and effect: can large language
  models truly understand causality?'' in \emph{Proceedings of the AAAI
  Symposium Series}, vol.~4, 2024, pp. 2--9.

\bibitem{carta2023grounding}
T.~Carta, C.~Romac, T.~Wolf, S.~Lamprier, O.~Sigaud, and P.-Y. Oudeyer,
  ``Grounding large language models in interactive environments with online
  reinforcement learning,'' in \emph{International Conference on Machine
  Learning}.\hskip 1em plus 0.5em minus 0.4em\relax PMLR, 2023, pp. 3676--3713.

\bibitem{chen2025next}
W.~Chen, H.~Huang, Z.~Zhang, T.~Wang, Y.~Lin, L.~Chang, and H.~Wan, ``Next-poi
  recommendation via spatial-temporal knowledge graph contrastive learning and
  trajectory prompt,'' \emph{IEEE Transactions on Knowledge and Data
  Engineering}, vol.~37, no.~6, pp. 3570--3582, 2025.

\bibitem{Dung2025Navigating}
D.~Nguyen, H.~Le, K.~Do, S.~Gupta, S.~Venkatesh, and T.~Tran, ``Navigating
  social dilemmas with llm-based agents via consideration of future
  consequences,'' in \emph{Proceedings of the Thirty-Fourth International Joint
  Conference on Artificial Intelligence, {IJCAI} 2025, Montreal, Canada, August
  16-22, 2025}, 2025, pp. 223--231.

\bibitem{gkountouras2024language}
J.~Gkountouras, M.~Lindemann, P.~Lippe, E.~Gavves, and I.~Titov, ``Language
  agents meet causality--bridging llms and causal world models,'' \emph{arXiv
  preprint arXiv:2410.19923}, 2024.

\bibitem{hao2025rl}
Q.~Hao, S.~Li, J.~Yuan, and Y.~Li, ``Rl of thoughts: Navigating llm reasoning
  with inference-time reinforcement learning,'' \emph{arXiv preprint
  arXiv:2505.14140}, 2025.

\bibitem{tang2025plan}
X.~Tang, M.~Kan, S.~Shan, and X.~Chen, ``Plan-r1: Safe and feasible trajectory
  planning as language modeling,'' \emph{arXiv preprint arXiv:2505.17659},
  2025.

\bibitem{gendron2025causal}
G.~Gendron, J.~M. Ro{\v{z}}anec, M.~Witbrock, and G.~Dobbie, ``Causal
  cartographer: From mapping to reasoning over counterfactual worlds,''
  \emph{arXiv preprint arXiv:2505.14396}, 2025.

\bibitem{dal2025joint}
L.~Dal'Col, M.~Oliveira, and V.~Santos, ``Joint perception and prediction for
  autonomous driving: A survey,'' \emph{IEEE Transactions on Intelligent
  Transportation Systems}, 2025.

\bibitem{ma2024coevolving}
H.~Ma, T.~Hu, Z.~Pu, L.~Boyin, X.~Ai, Y.~Liang, and M.~Chen, ``Coevolving with
  the other you: Fine-tuning llm with sequential cooperative multi-agent
  reinforcement learning,'' in \emph{Proceedings of Neural Information
  Processing Systems}, vol.~37, 2024, pp. 15\,497--15\,525.

\bibitem{zhai2024fine}
S.~Zhai, H.~Bai, Z.~Lin, J.~Pan, P.~Tong, Y.~Zhou, A.~Suhr, S.~Xie, Y.~LeCun,
  Y.~Ma \emph{et~al.}, ``Fine-tuning large vision-language models as
  decision-making agents via reinforcement learning,'' \emph{Advances in neural
  information processing systems}, vol.~37, pp. 110\,935--110\,971, 2024.

\bibitem{requeima2024llm}
J.~Requeima, J.~Bronskill, D.~Choi, R.~Turner, and D.~K. Duvenaud, ``Llm
  processes: Numerical predictive distributions conditioned on natural
  language,'' in \emph{Proceedings of Neural Information Processing Systems},
  vol.~37, 2024, pp. 109\,609--109\,671.

\bibitem{lin2025improving}
X.~Lin, Y.~Wu, H.~Yang, Y.~Huang, Y.~Zhang, J.~Ji, and Y.~Zhang, ``Improving
  efficiency of answer set planning with rough solutions from large language
  models for robotic task planning,'' in \emph{Proceedings of the Thirty-Fourth
  International Joint Conference on Artificial Intelligence}, 2025, pp.
  4570--4578.

\bibitem{song2023llm}
C.~H. Song, J.~Wu, C.~Washington, B.~M. Sadler, W.-L. Chao, and Y.~Su,
  ``Llm-planner: Few-shot grounded planning for embodied agents with large
  language models,'' in \emph{Proceedings of the IEEE/CVF international
  conference on computer vision}, 2023, pp. 2998--3009.

\bibitem{Ahn2022can}
M.~Ahn, A.~Brohan, N.~Brown, Y.~Chebotar, O.~Cortes, B.~David, C.~Finn, C.~Fu,
  K.~Gopalakrishnan, and K.~Hausman, ``Do as i can, not as i say: Grounding
  language in robotic affordances,'' \emph{arXiv preprint arXiv:2204.01691},
  2022.

\bibitem{shinn2023reflexion}
N.~Shinn, F.~Cassano, A.~Gopinath, K.~Narasimhan, and S.~Yao, ``Reflexion:
  Language agents with verbal reinforcement learning,'' in \emph{Proceedings of
  Neural Information Processing Systems}, vol.~36, 2023, pp. 8634--8652.

\bibitem{liang2023code}
J.~Liang, W.~Huang, F.~Xia, P.~Xu, K.~Hausman, B.~Ichter, P.~Florence, and
  A.~Zeng, ``Code as policies: Language model programs for embodied control,''
  in \emph{Proceedings of IEEE International Conference on Robotics and
  Automation}, 2023, pp. 9493--9500.

\bibitem{hazra2024saycanpay}
R.~Hazra, M.~Dos, Z.~Pedro, and L.~De~Raedt, ``Saycanpay: Heuristic planning
  with large language models using learnable domain knowledge,'' in
  \emph{Proceedings of the AAAI Conference on Artificial Intelligence},
  vol.~38, 2024, pp. 20\,123--20\,133.

\bibitem{kambhampati2024position}
S.~Kambhampati, K.~Valmeekam, L.~Guan, M.~Verma, K.~Stechly, S.~Bhambri, L.~P.
  Saldyt, and A.~B. Murthy, ``Position: Llms can’t plan, but can help
  planning in llm‑modulo frameworks,'' in \emph{Proceedings of the
  International Conference on Machine Learning}, vol. 235, 2024, pp.
  22\,895--22\,907.

\bibitem{liu2025diffugc}
B.~Liu, H.~Li, and S.~Hong, ``{DiffuGC}: Diffusion model can help discover
  granger causality from interventional time series,'' in \emph{IEEE
  International Conference on Data Mining (ICDM)}.\hskip 1em plus 0.5em minus
  0.4em\relax IEEE, 2025, pp. 487--496.

\bibitem{zhou2021multiobjrl}
F.~Zhou, C.~Lu, X.~Tang, F.~Zhang, Z.~Qin, J.~Ye, and H.~Zhu, ``Multi-objective
  distributional reinforcement learning for large-scale order dispatching,'' in
  \emph{IEEE International Conference on Data Mining (ICDM)}.\hskip 1em plus
  0.5em minus 0.4em\relax IEEE, 2021, pp. 1541--1546.

\bibitem{qian2024adaptraj}
T.~Qian, Y.~Chen, G.~Cong, Y.~Xu, and F.~Wang, ``Adaptraj: A multi-source
  domain generalization framework for multi-agent trajectory prediction,'' in
  \emph{2024 IEEE 40th International Conference on Data Engineering
  (ICDE)}.\hskip 1em plus 0.5em minus 0.4em\relax IEEE, 2024, pp. 5048--5060.

\bibitem{hu2025efficient}
J.~Hu, Y.~Wang, S.~Zhang, K.~Zhou, G.~Chen, Y.~Hu, B.~Xiao, and M.~Tan,
  ``Efficient dynamic ensembling for multiple llm experts,'' in
  \emph{Proceedings of the Thirty-Fourth International Joint Conference on
  Artificial Intelligence, IJCAI}, 2025, pp. 16--22.

\bibitem{zhu2023towards}
T.~Zhu, Y.~Qiu, H.~Zhou, and J.~Li, ``Towards long-delayed sparsity: Learning a
  better transformer through reward redistribution.'' in \emph{Proceedings of
  the Thirty-Second International Joint Conference on Artificial Intelligence},
  2023, pp. 4693--4701.

\bibitem{liao2024cdstraj}
H.~Liao, X.~Li, Y.~Li, H.~Kong, C.~Wang, B.~Wang, Y.~Guan, K.~Tam, and Z.~Li,
  ``Cdstraj: Characterized diffusion and spatial-temporal interaction network
  for trajectory prediction in autonomous driving,'' in \emph{Proceedings of
  the Thirty-Third International Joint Conference on Artificial Intelligence,
  IJCAI-24}, 2024, pp. 7331--7339.

\bibitem{ha2023scaling}
H.~Ha, P.~Florence, and S.~Song, ``Scaling up and distilling down:
  Language-guided robot skill acquisition,'' in \emph{Conference on Robot
  Learning}.\hskip 1em plus 0.5em minus 0.4em\relax PMLR, 2023, pp. 3766--3777.

\bibitem{highway-env}
E.~Leurent, ``An environment for autonomous driving decision-making,''
  \url{https://github.com/eleurent/highway-env}, 2018.

\end{thebibliography}

\clearpage
\appendices
\section{Implementation Details}
\label{sec:Appendix_A}
Table~\ref{tab:impl_details} summarizes the main implementation details. For exact definitions and usage of each parameter, please refer to our released anonymized source code.

\begin{table}[htbp]
\setlength{\tabcolsep}{4.5pt}
\centering
\caption{Implementation Details.}
\label{tab:impl_details}
\renewcommand{\arraystretch}{1.19}
\begin{tabular}{p{3.cm} p{4.5cm}}
\toprule
\multicolumn{2}{c}{\textbf{Environment Configuration}} \\
\midrule
Frequency & 30 \\
Lanes count & \{2, 3, 4\} \\
Vehicles density & \{2, 3\} \\
\midrule
\multicolumn{2}{c}{\textbf{LLM Module}} \\
\midrule
Model & GPT-4.1-nano / GPT-3.5-turbo \\
Action Space & DiscreteMetaAction (5 actions) \\
Rollout Length $L$ & \{8, 10, 12, 16\} \\
Temperature & 0.7 \\
\midrule
\multicolumn{2}{c}{\textbf{Grounder Module}} \\
\midrule
Model & Multi-headed Transformer \\
Warm-up Samples & \{0, 8k, 12k, 18k, 24k\} \\
State Dimension & $N \times 4$ \\
Optimizer / LR & Adam / $1 \times 10^{-4}$ \\
\midrule
\multicolumn{2}{c}{\textbf{Distiller Module}} \\
\midrule
Model & 3-layer MLP, 32 hidden units \\
Distiller Weight $\alpha$ & 0.9 \\
Utility Weights $(\omega_P,\omega_E,\omega_C)$ & (20, 0.3, 0.2) \\
Optimizer / LR & Adam / $3 \times 10^{-3}$ \\
\midrule
\multicolumn{2}{c}{\textbf{Policy Module}} \\
\midrule
Algorithm & Transformer + PPO \\
Sequence Length & 4 \\
Discount Factor $\gamma$ & 0.95 \\
GAE $\lambda$ & 0.95 \\
PPO Clip $\epsilon$ & 0.2 \\
Entropy Coefficient & 0.1 \\
Optimizer / LR & Adam / $3 \times 10^{-4}$ \\
Training Setup & 4 parallel envs, 1000 steps \\
\bottomrule
\end{tabular}
\end{table}

\section{Structural Multi-dimensional Effects Prompt}
\label{sec:Appendix_B}
To ensure transparency and reproducibility, we provide the full structured multi-dimensional effect prompt used by the LLM-based Effect Evaluator in Fig.~\ref{fig:llmpromptbox}. For each candidate action, the LLM is instructed to produce four effect estimates, such as reward, crash risk, energy, and comfort, together with a short textual explanation that serves only as a reference for the numerical values. To improve robustness under dynamic conditions, we query the LLM three times for each online data frame and average the returned estimates; across repeated queries the estimates remain stable with only minor numerical variations, supporting the LLM's credibility as a reasoning module in the planning loop. Representative raw responses are released in our anonymized code repository.

\begin{figure*}[h!]
\centering
\begin{tcolorbox}[
  colback=gray!8, colframe=black!80,
  boxrule=0.5pt, arc=3mm,
  left=8pt, right=8pt, top=6pt, bottom=6pt,   
  title=Structural Multi-dimensional Effects Prompt,
  colbacktitle=gray!40, coltitle=black, fonttitle=\bfseries
]
\small
\textbf{[Task]}

You are an expert in autonomous driving trajectory planning.  Your task is to generate \textbf{structured, multi-dimensional effect estimates} for each candidate action, following the four steps below.

\textbf{(A) Contextual Observation.}

You are given a sample of online environment data: {\footnotesize\ttfamily \{online data\}},

Please carefully observe the current environment and summarize key planning relevant factors such as:
\begin{itemize}
    \item  ego vehicle position, lane, and velocity,
    \item relative positions and velocities of nearby vehicles,
    \item road structure and available action options.
\end{itemize}

This information will be used to support subsequent action-conditioned effect evaluation.

\textbf{(B) Action-Conditioned Rollout Evaluation.}

Action-conditioned rollout evaluation defines the potential consequences of executing an action at the current state over the rollout horizon.

You are given a discrete action set:\{“IDLE”, “Turn-Right”, “Turn-Left”, “Acceleration”, “Deceleration”\}

For each action $a$, perform action-conditioned rollout evaluation to estimate the following multi-dimensional effect definitions, explicitly considering key planning-relevant factors such as ego dynamics, surrounding vehicle interactions, and road structure.

\textbf{(C) Effect-grounded Semantic Definition. }

For each action, report the following four effects over the \textbf{rollout horizon}:
\begin{itemize}
    \item \textbf{Reward} (float):Reward is the expected cumulative reward over the rollout horizon after taking action a. It reflects the benefit of the action in terms of progress and rule compliance. Positive reward for safe maneuver completion and forward movement; penalty for lane departure or traffic violations.

\item \textbf{Crash} (float in [0, 1]): Crash is the expected probability of collision over the rollout horizon after taking action \textit{a}. It captures the risk of hitting any object during execution.

\item \textbf{Energy} (float): Energy is the expected energy consumption over the rollout horizon after action \textit{a}, approximated as the sum of squared acceleration magnitudes.

\item \textbf{Comfort} (float): Comfort is the expected driving comfort over the rollout horizon after action \textit{a}, measured as the negative sum of acceleration change and abrupt directional shifts.

\end{itemize}

\textbf{(D)  Structured Quantitative Output.}

Please return \textbf{EXACTLY} a JSON array of objects, sorted by descending reward.  

Each object must include the following fields:

\{"Action":"string",  "Reward": float, "Crash": float, "Energy": float, "Comfort": float, "Explanation":  "string" \}

\medskip
\begin{verbatim}
All numeric fields must be numeric literals (e.g.,2.34), not words." 
\end{verbatim}

\end{tcolorbox}
\caption{The prompt of structural multi-dimensional effects.}
\label{fig:llmpromptbox}

\end{figure*}

\section{Computational Overhead Analysis}
\label{sec:Appendix_E}
We evaluate the computational efficiency of \textsf{GAPL} against five LLM-based baselines, all using GPT-4.1-nano as the unified backbone on the same hardware (NVIDIA A10 GPU) and identical environment configurations.

\begin{table}[htbp]
\centering
\caption{Computational Overhead Comparison. ``Time'' denotes system inference time per decision step (ms). ``Calls'' denotes the number of LLM API calls per step.}
\label{tab:computational_overhead}
\setlength{\tabcolsep}{1.2pt}
{
\fontsize{8.4pt}{10pt}\selectfont
\begin{tabular}{c|c|cccccc}
\toprule
& & SayCan & Reflexion & Plan-R1 & Cartographer & LLMCWM & \textbf{GAPL} \\
\midrule
\multirow{2}{*}{\textit{Rou.}} &
Time & 756 & 1124 & 612 & 934 & 823 & \textbf{125} \\
& Calls & 6.2 & 9.4 & 5.1 & 7.8 & 6.9 & \textbf{1.0} \\
\midrule
\multirow{2}{*}{\textit{Int.}} &
Time & 893 & 1387 & 743 & 1089 & 967 & \textbf{138} \\
& Calls & 7.4 & 11.6 & 6.2 & 9.1 & 8.1 & \textbf{1.0} \\
\bottomrule
\end{tabular}
}
\end{table}

As shown in Table~\ref{tab:computational_overhead}, \textsf{GAPL} achieves substantially lower latency and far fewer LLM calls than all baselines. It requires only a single API call per decision step (an 0.804--0.914 reduction over the 5--12 calls of baselines) and reduces inference time to 125--138\,ms (a 0.80--0.90 reduction over the fastest and slowest baselines, respectively). The efficiency gain stems from our policy-based control design: by offloading numerical control to a lightweight PPO policy with a 13\,ms grounder, \textsf{GAPL} avoids repeated LLM queries for action refinement. The resulting 7--8\,Hz decision frequency, together with the single-call design that lowers per-episode API cost by 5--12$\times$, makes \textsf{GAPL} a practical candidate for real-time autonomous driving.

\section{Sensitivity Study on Weight Configurations}
\label{sec:Appendix_F}
As defined in Eq.~(\ref{equ:shapedreward}), the shaped reward is controlled by three hyperparameters, $\omega_{\text{LLM}}$, $\omega_{\text{GRO}}$, and $\omega_{\text{env}}$, under the constraint $\omega_{\text{LLM}}+\omega_{\text{GRO}}+\omega_{\text{env}}=1$. We study how different allocations affect reward learning by evaluating four representative schemes, where each triple denotes $(\omega_{\text{LLM}},\omega_{\text{GRO}},\omega_{\text{env}})$: Red $(0.3, 0.1, 0.6)$, Green $(0.5, 0.2, 0.3)$, Yellow $(0.2, 0.6, 0.2)$, and Blue $(0.3, 0.6, 0.1)$.

\begin{figure*}[htbp]
    \centering
    \includegraphics[width=\linewidth]{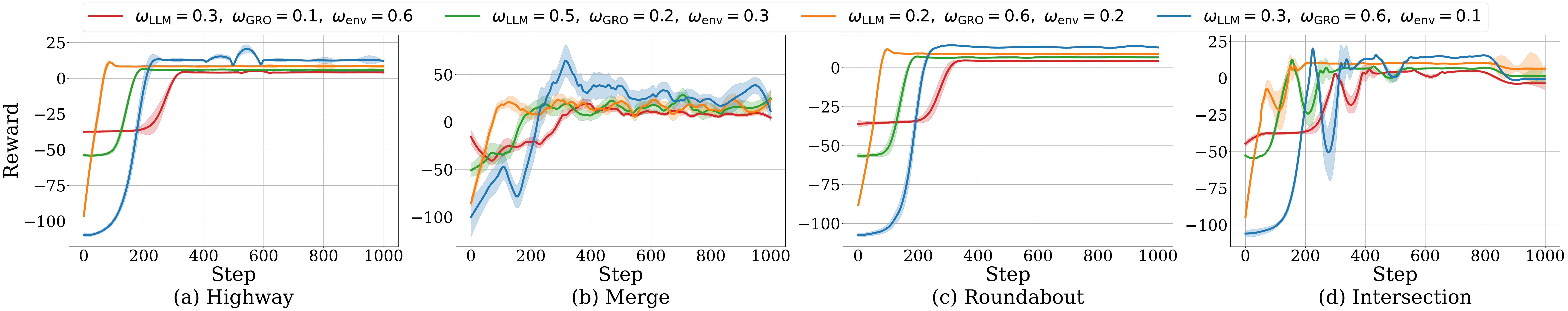}
    \caption{Training curves of reward under varying weighting configurations across environments.}
    \label{fig:sensitiveCurve}
\end{figure*}

As shown in Fig.~\ref{fig:sensitiveCurve}, the Blue configuration consistently achieves the highest final reward across all environments. By assigning the dominant weight to the simulation-based grounder while retaining a moderate LLM contribution and the smallest environment weight, it effectively integrates language priors with dynamics-consistent effect estimation, yielding more stable and coherent planning, particularly in highly interactive scenarios such as \emph{Roundabout} and \emph{Intersection}. In contrast, environment-dominated schemes (e.g., Red) exhibit rapid early gains but converge to lower rewards, while LLM-dominated schemes without sufficient grounding also converge suboptimally. These results confirm that jointly leveraging semantic reasoning, simulation-grounded effects, and environment feedback is essential for stable trajectory planning.

\section{Sensitivity Study on LLM Backbones}
\label{sec:Appendix_LLMbb}
To examine how planning performance depends on the capacity of the LLM-based effect estimator, we replace the LLM module with different backbones and evaluate under the two most complex interactive scenarios, \emph{Roundabout} and \emph{Intersection}.

As shown in Table~\ref{tab:generalization}, GPT-4.1-nano consistently outperforms GPT-3.5-turbo across all metrics, reducing ADE by 0.56/0.86, FDE by 0.33/1.13, and increasing reward by 0.55/0.64 in the two scenarios. This indicates that stronger language priors yield more accurate effect estimation. Nonetheless, the relatively small gaps suggest that lightweight backbones remain competitive, highlighting the potential for efficient effect evaluators in complex multi-agent environments.

\begin{table}[htbp]
\centering
\caption{Trajectory Planning Performance With Different LLM Backbones in the Roundabout and Intersection Scenarios.}
\label{tab:generalization}
\setlength{\tabcolsep}{1.2pt}
{
\fontsize{8.4pt}{10pt}\selectfont
\begin{tabular}{c|c|ccc}
\toprule
& \multicolumn{1}{c|}{Model} & \textbf{ADE} ($\downarrow$)& \textbf{FDE} ($\downarrow$) & \textbf{Reward} ($\uparrow$) \\
\midrule
\multirow{2}{*}{Rou.} & GPT-3.5-turbo   & 7.78 ± 1.11 & 9.46 ± 1.19 & 6.29 ± 0.37 \\
&  GPT-4.1-nano      & \textbf{7.22 ± 1.04} & \textbf{9.13 ± 1.06} & \textbf{6.84 ± 0.41} \\
\midrule
\multirow{2}{*}{Int.}&  GPT-3.5-turbo     & 10.38 ± 1.22 & 12.36 ± 1.25 & 5.69 ± 0.77 \\
&  GPT-4.1-nano      & \textbf{9.52 ± 0.98} & \textbf{11.23 ± 1.01} & \textbf{6.33 ± 0.46} \\
\bottomrule
\end{tabular}
}
\end{table}

\end{document}